\documentclass[runningheads]{llncs}
\usepackage{graphicx}
\usepackage[colorlinks]{hyperref}
\usepackage{graphicx}
\usepackage{amsmath}
\usepackage{amssymb}
\usepackage{times}
\usepackage{epsfig}
\usepackage{graphicx}
\usepackage{float}
\usepackage{todonotes}
\usepackage{tabularx}

\begin{document}
%\newcolumntype{L}[1]{>{\raggedright\arraybackslash}p{#1}}
\newcolumntype{C}[1]{>{\centering\arraybackslash}p{#1}}
%\newcolumntype{R}[1]{>{\raggedleft\arraybackslash}p{#1}}

\title{3D Organ Shape Reconstruction from Topogram Images}
\author{Elena Balashova\inst{1,2} \and 
Jiangping Wang\inst{2} \and 
Vivek Singh\inst{2} \and 
Bogdan Georgescu\inst{2} \and 
Brian Teixeira\inst{2} \and 
Ankur Kapoor\inst{2}}
% \author{Anonymous Submission \# 96}

\authorrunning{E. Balashova et al.}
\institute{Department of Computer Science, Princeton University, Princeton, NJ, USA
\and
Siemens Healthineers, Digital Services, Digital Technology \& Innovation, Princeton, NJ, USA
}

% \authorrunning{Anonymous Submission \# 96}

\maketitle          
\begin{abstract}
Automatic delineation and measurement of main organs such as liver is one of the critical steps for assessment of hepatic diseases, planning and postoperative or treatment follow-up. However, addressing this problem typically requires performing computed tomography (CT) scanning and complicated post-processing of the resulting scans using slice-by-slice techniques. In this paper, we show that 3D organ shape can be automatically predicted directly from topogram images, which are easier to acquire and have limited exposure to radiation during acquisition, compared to CT scans. We evaluate our approach on the challenging task of predicting liver shape using a generative model. We also demonstrate that our method can be combined with user annotations, such as a 2D mask, for improved prediction accuracy. We show compelling results on 3D liver shape reconstruction and volume estimation on $2129$ CT scans. 
\footnote{This feature is based on research, and is not commercially available. Due to regulatory reasons its future availability cannot be guaranteed.} 
\end{abstract}

\section{Introduction}
% goal
In medical imaging, observing realistic organ shape is a critical step in enabling health care professionals gain a better insight into patients' body. Accurate depiction of internal organs, such as liver, often allows for more accurate health screening and early diagnosis, as well as planning of procedures such as radiation therapy to target specific locations in the human body. Delineating 3D organ shape from 2D X-ray images is an extremely difficult and unsolved problem in bio-medical engineering today due to visual ambiguities and information loss as a result of projection. The goal of this problem is to accurately predict the shape of the observed 3D organ given a single image.

% challenge
Existing liver delineation techniques typically produce organ shape from computed tomography (CT) scans. The procedures to obtain these scans involve long patient-doctor interaction time, costly machinery, and exposure to a high dose of radiation. The practical challenges in obtaining these scans may preclude obtaining accurate organ depictions. In addition, existing delineation tools~\cite{yang2017automatic} would delineate (either automatically or semi-automatically) the two-dimensional shape in each slice of the three-dimensional CT volume and combine the set of predictions into a three-dimensional shape. The intermediate processing may introduce an additional source of error to the overall shape prediction quality due to the lack of spatial context.

% approach
The key idea of this paper is to reconstruct 3D organ shape from topograms, which are projected 2D images from tomographic devices, such as X-ray~\cite{sioutos2007nci}. These types of images can be much more easily obtained and are often used by medical professionals for planning purposes~\cite{mayo2014managing,schertler2007dual}. Motivated by the significant advances in deep learning techniques for organ segmentation~\cite{Zhang2018TaskDG} and representation learning on 3D data ~\cite{girdhar2016learning,sharma2016vconv,qi2016volumetric}, we pose the problem of organ reconstruction as the task of predicting 3D shape from a single image. Further, we describe an automatic delineation procedure that outputs the shape from the topogram image only, as well as a semi-automatic extension, where we allow the user to outline the approximate two-dimensional mask and use it (in conjunction with the topogram) to obtain a more accurate 3D shape prediction.

% overview
Our system has two components: a generative shape model, composed of a shape encoder and decoder, and an encoder from 2D observations (topogram only or topogram and mask). The shape encoder and decoder form a variational auto-encoder (VAE)~\cite{kingma2014vae} generative model in order to represent each shape observation using a compact low-dimensional representation. The topogram and optional mask encoders (whose architectures are similar to \cite{wu3dGAN}) map the partial observations from images (and masks when provided) to the coordinates of the corresponding shape observations. The entire system is optimized end-to-end in order to simultaneously infer shapes from topogram image observations and to learn the underlying shape space. This allows us to simultaneously learn a generative shape space covering complex shape variations from the 3D supervisions and infer the shapes from input 2D observations. To validate our approach, we collected a new medical dataset of $2129$ abdominal CT scans and topogram images, and evaluated the proposed approach on the challenging tasks of 3D liver reconstruction and volume prediction. The contributions of our work are:
\begin{itemize}
    \item An automatic and a semi-automatic approach to perform 3D organ reconstruction from 2D topograms, allowing automatic 3D shape prediction from the topogram only and a more refined prediction where 2D mask annotation is available.
    \item An evaluation of our method on accurate 3D organ volume estimation and reconstruction applications.
\end{itemize}

\section{Related Work}
% liver segmentation
In the medical imaging domain, extraction and visualization of 3D organs is a key step in clinical applications such as surgical planning and post-surgical assessment, as well as pathology detection and disease diagnosis. Of particular interest is the liver, which can exhibit highly heterogeneous shape variation that makes it even more difficult to segment. Previously, liver volume was segmented semi-automatically~\cite{hame2012semi} or automatically using statistical shape models~\cite{heimann2009comparison}, sigmoid-edge modelling~\cite{foruzan2016improved}, graph-cut~\cite{li2015automatic} and others (see \cite{mharib2012survey} for an overview). Recently, automatic deep learning based methods~\cite{christ2017automatic,dou20163d,lu2017automatic} have been shown to provide impressive results on this task. However, these methods need a CT scan procedure, which is costly and requires a high radiation exposure. On the other hand, X-ray and topogram images are easier to obtain, require less radiation, and are often used by medical professionals for planning purposes~\cite{mayo2014managing,schertler2007dual}. 

Shape extraction from X-ray is particularly complex as its projective nature can contain complex or fuzzy textures, boundaries and anatomical part overlap~\cite{Zhang2018TaskDG}. To mitigate these challenges, traditional methods use prior knowledge, such as motion patterns~\cite{zhu2009dynamic} or intensity and background analysis~\cite{qin2019accurate}, in order to perform X-ray segmentation. More recent methods~\cite{ronneberger2015u} focus on learning to segment using deep neural networks. For example, \cite{albarqouni2017x} decomposes X-ray into non-overlapping components, ~\cite{yang2017automatic} uses a generative adversarial network (GAN)~\cite{wu3dGAN} to improve segmentation quality, and ~\cite{Zhang2018TaskDG} applies unpaired image-image translation techniques to learn to segment X-ray by observing CT scan segmentation. These methods achieve remarkable results on 2D shape delineation and segmentation tasks. 

In parallel, in the computer vision domain, deep generative 3D shape models based on variational auto-encoder networks (VAE)~\cite{girdhar2016learning,sharma2016vconv} and generative adversarial networks (GAN)~\cite{wu3dGAN} have shown superior performance in learning to generate complex topologies of shapes. Combined with a mapping from image space, these methods are able to infer 3D shape predictions from 2D observations. To obtain more detailed and accurate predictions, input annotations, such as landmarks or masks, are often used to guide the synthesis process. \cite{bogo2016keep} incorporates 2D landmarks for alignment optimization of a skinned vertex-based human shape model to image observations. \cite{kar2015category} and \cite{vicente2014reconstructing} applies landmark annotations to guide synthesis of observed 3D shape in input images. \cite{balashova2018structure} uses landmarks and \cite{gadelha20173d} incorporates silhouettes to formulate additional objective terms to improve performance in 3D shape reconstruction and synthesis problems. 

To the best of our knowledge, we are the first to propose both automatic and semi-automatic approaches to 3D organ shape reconstruction from topograms.

\section{Overview}
\begin{figure}[!ht]
    \centering
    \begin{minipage}{0.95\linewidth}
        \centering
        \includegraphics[width=\linewidth]{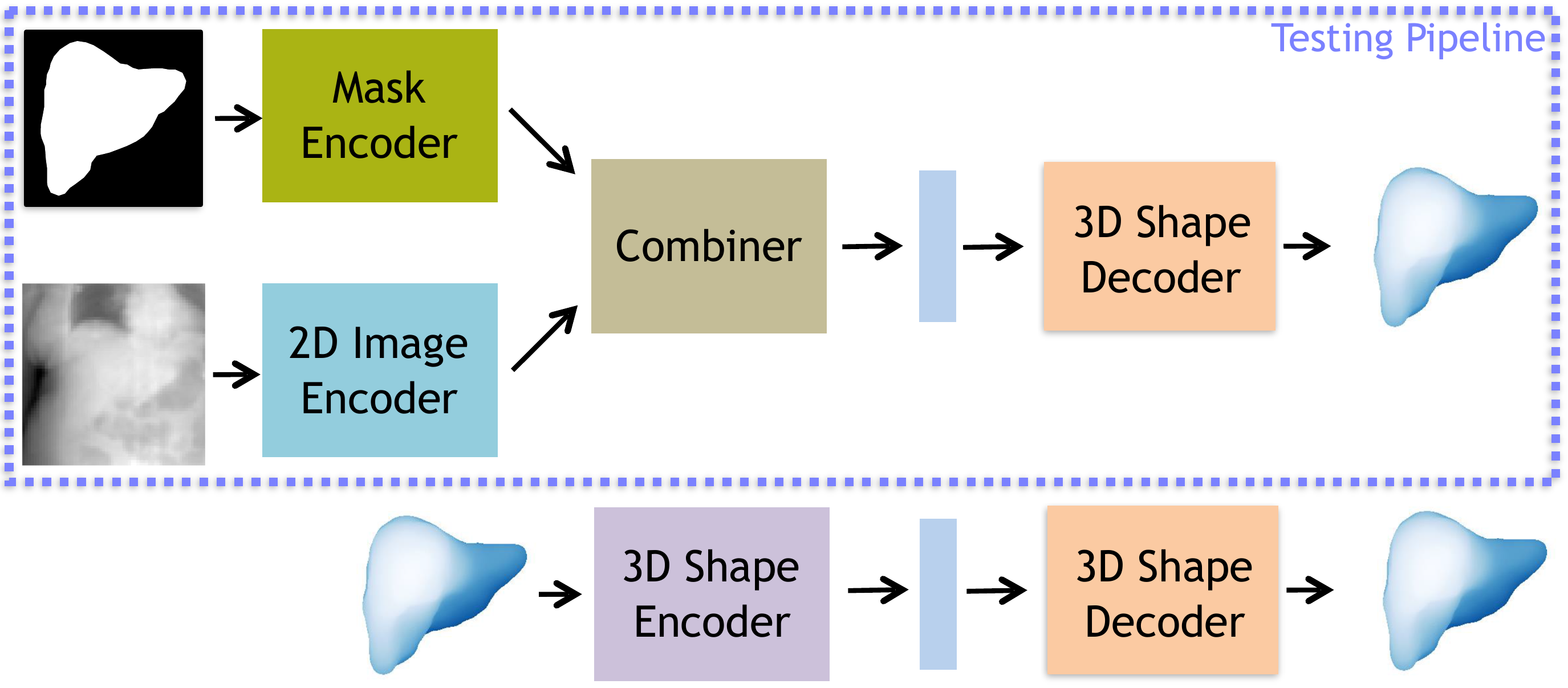}\\
        (a) 2D-3D organ delineation from topograms and provided masks. 
        % \label{fig:topo-mask}
    \end{minipage}
    % \vspace{1.5cm}
    \begin{minipage}{0.77\linewidth}
        \centering
        \includegraphics[width=\linewidth]{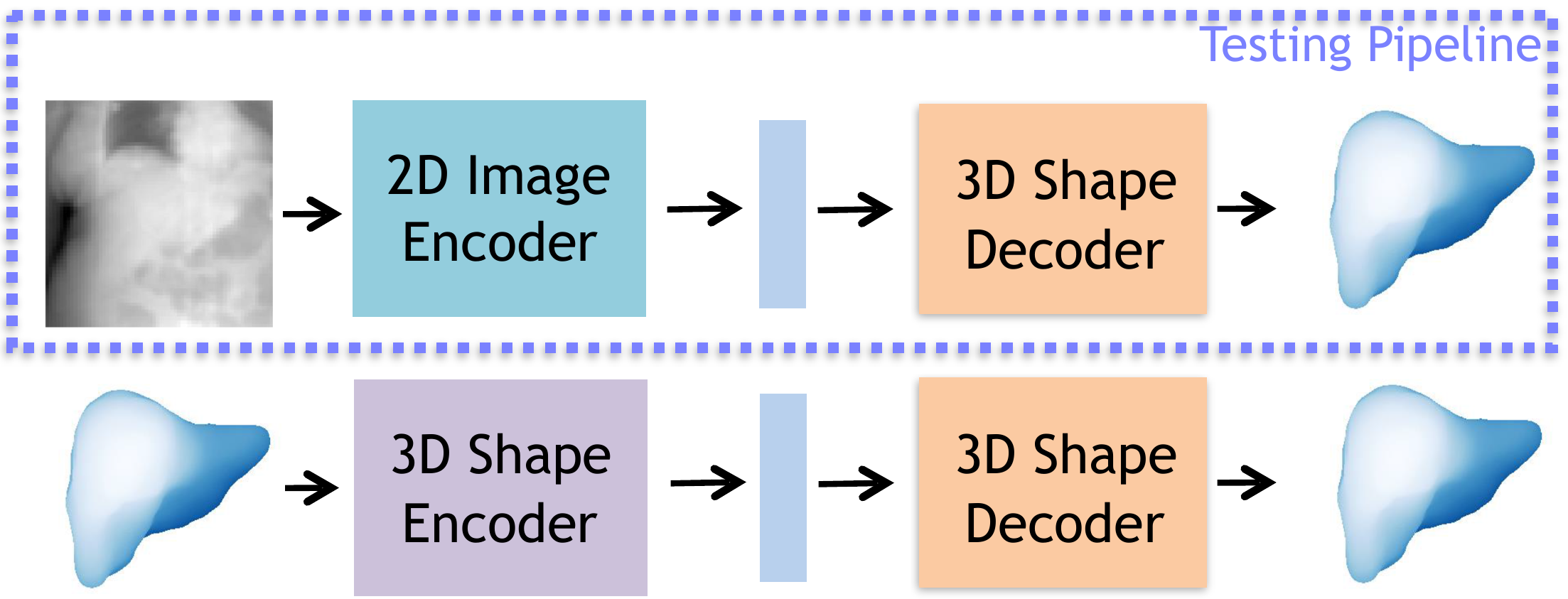}\\
        (b) 2D-3D organ delineation from topograms only.
        % \label{fig:topo-only}
    \end{minipage}
        \caption{Overview of our system. We train a generative model from a collection of 3D shapes, and learn to map from topograms  (see Figure~\ref{fig:overview}$(b)$) or topograms and user-provided 2D masks (see Figure~\ref{fig:overview}$(a)$) to reconstructed 3D shapes of the observed organ. For both methods, the training phase involves training the generative model (3D shape encoder and decoder) jointly with the 2D observation encoders (topogram (blue), or topogram (blue), mask (green) and their combiner (brown)) in an end-to-end procedure described in  Section~\ref{sec:combinedTraining}. During testing, only the 2D observations are necessary for 3D shape prediction. }
    \label{fig:overview}
\end{figure}

An overview of our training pipeline can be seen in Figure~\ref{fig:overview}. Our system consists of several key components: a generative shape model and a set of encoders from 2D observations. The generative model is composed of an encoder and a decoder, where the encoder maps the 3D shapes of organs to their coordinates in the latent space and the decoder reconstructs the shapes back from their coordinates. The first observation encoder is the topogram encoder that maps two-dimensional observations to the coordinates of the corresponding shapes. The second observation encoder is the joint topogram and mask encoder that predicts the latent coordinate of the organ shape given the 2D mask and topogram. The mask information, when provided, helps generate a more accurate prediction. 

The organ shape prediction approach is very general, and can be used for organs other than human liver. The technique requires access to a database of shape and  X-ray (two-dimensional observation) pairs. We demonstrate an accuracy improvement using user input in the form of 2D masks. Other types of input that can be encoded using a neural network can also be applied in place of masks to improve prediction accuracy.

\subsubsection{Generative Model}
As input, our system receives a set of examples $E = \big\{(s,i)\big\}$ where $s\in S$ is the example shape and $i\in I$ is the corresponding topogram image observation. The generative model $G = (Q,P)$ consists of an encoding component $Q$ and a decoding component $P$. Here $Q(z|s)$ maps shape $s$ to its latent coordinate $z$ in the stochastic low dimensional space distributed according to prior distribution $p(z)$ and $P(s|z)$ maps the latent coordinate $z$ back to the shape space $S$. The loss function of the generative model is composed of a reconstruction loss $L_{rec}$ and a distribution loss $L_{dist}$, as is typical for variational auto-encoder training. $L_{rec}$ is the binary cross entropy (BCE) error that measures the difference between the ground truth shape $s \in S$ and the predicted shape $s'\in S$:
\begin{eqnarray} \label{eq:bce}
L_{rec}(s,s') = -\frac{1}{N}\sum^{N}_{n=1} s_n \log{s'_n} + (1-s_n)\log{(1-s'_n)} 
\end{eqnarray}
where $N=64^3$. $L_{dist}$ is the distribution loss that enforces the latent distribution of $z_1$ to match its prior distribution $L_{dist}(z_1) =  \mathrm{KL}\left(Q(z|s) \| p(z)\right)$, where $p(z)= \mathcal{N}(\mu,\,\sigma^{2})$ and ${\alpha}_1,{\alpha}_2$ are the weights applied to each type of loss.

The 3D shape encoder maps an observation, represented with a $64$ by $64$ by $64$ voxel grid, to its $200$-dimensional latent vector $z$. The normal distribution parameters are defined $\mu=0$ and $\sigma = 1$, as is customary for variational auto-encoder models.  The architecture of the encoder consists of five convolutional layers with output sizes $64,128,256,512,200$, kernel size $4$ for each layer, and padding sizes $1$,$1$,$1$,$1$ and $0$. The convolutional layers are separated by batch-normalization~\cite{batchnorm} and ReLU layers~\cite{relu}. The 3D shape decoder takes as input a single $200$-dimensional latent vector $z$, and predicts a $64$ by $64$ by $64$ voxelized representation of shape. The decoder architecture mirrors that of the encoder.

\subsubsection{Topogram Encoder} \label{sec:topo_enc}
Given a generative model $G$, we can learn a topogram image encoder $I_1$, so that for each observation $(s,i)\in E$, the image $i$ is mapped to the coordinate location $\hat{z} = I_1(i)$ such that the reconstructed shape $G(\hat{z})$ and the ground truth shape $s$ are as close as possible. The image encoder loss is the binary cross entropy (BCE) loss $L_{rec}(s,G(\hat{z}))$ as defined in Equation~\ref{eq:bce}.  

The topogram encoder $I_1$ takes a $1$ by $256$ by $256$ topogram image, and outputs a $200$-dimensional latent shape vector $\hat{z}$. It consists of five convolutional layers with the number of outputs $64,128,256,512,200$, kernel sizes $11,5,5,5,8$ and strides $4,2,2,2,1$, separated by batch-normalization~\cite{batchnorm} and rectified linear units (ReLU)~\cite{relu}.

\subsubsection{Topogram and Mask Encoder} \label{sec:topoCont}
For each observation $(s,i)\in E$, given a topogram $i$ and a mask $k=Pr(s)\in K$, where $Pr(\cdot)$ is defined to be an orthographic projection operator, we train the joint topogram and mask encoder $I_2$ that outputs $\tilde{z}=I_2(i,k)$ so that $G(\tilde{z})$ and $s$ are as close as possible. The loss of $I_2$ is defined to be the binary cross entropy (BCE) error $L_{rec}(s,G(\tilde{z}))$, as defined Equation~\ref{eq:bce}. We also enforce an additional mask loss:
\begin{eqnarray*}
L_{mask}(k,\tilde{k}) = -\sum^{N}_{n=1} k_n \log{\tilde{k}_n} + (1-k_n)\log{(1-\tilde{k}_n)}.
\end{eqnarray*}
that ensures that the input mask $k$ and the projected mask $\tilde{k}$ of the predicted shape (i.e. $\tilde{k}=Pr(G(\tilde{z}))$) match.

The topogram and mask encoder $I_2$ consists of a topogram encoder branch, a mask encoder branch, and a common combiner network (see Figure~\ref{fig:overview}), so that the observations are mapped to a common latent coordinate $\tilde{z}$. The topogram encoder branch has the same architecture as the topogram encoder in Section~\ref{sec:topo_enc} and maps the topogram to an intermediate $200$-dimensional feature $v_1$. The mask encoder branch receives a $1$ by $64$ by $64$ binary mask image which it maps to a $200$-dimensional vector $v_2$ using five convolutional layers with kernel sizes of $3,3,3,3,3$ and strides $4,2,2,2,2$, separated by batch-normalizations~\cite{batchnorm} and rectified linear units (ReLU)~\cite{relu}. $v_1$ and $v_2$ are then concatenated and run through the the combiner network consisting of a single fully connected layer to predict a joint $200$-dimensional latent coordinate $\tilde{z}$.

\begin{figure*}[!t]
\centering
\includegraphics[width=0.9\linewidth]{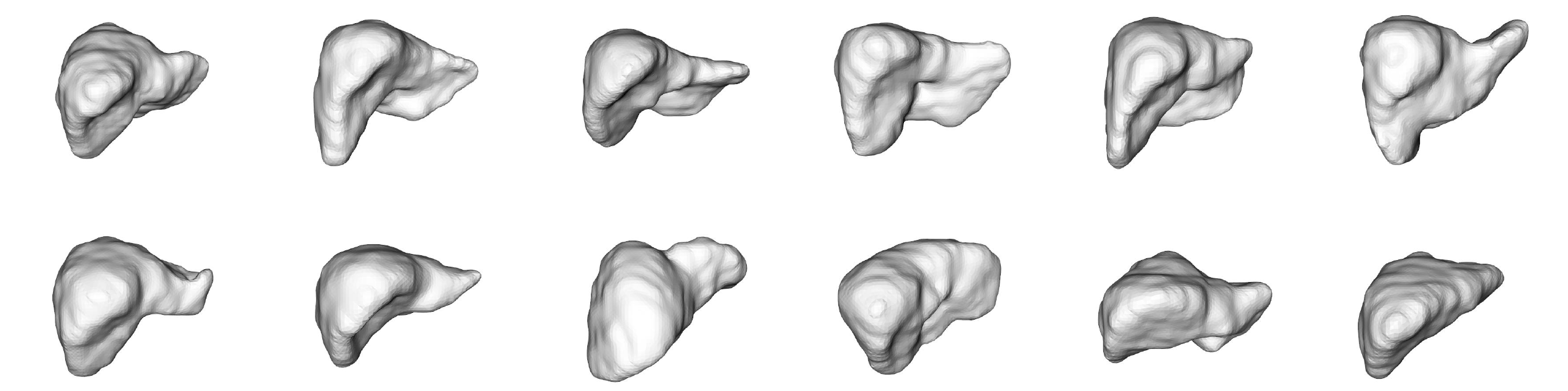}
\caption{Visualization of example 3D shape variations of the liver in the collected medical dataset of abdominal CT scans. The shapes represent a complex assortment typical of this organ.}
\label{fig:examples-liver}
\end{figure*}

\subsubsection{Combined Training } \label{sec:combinedTraining}
To train the models, we optimize the all the components of the system together in an end-to-end training process using the combined objective:
\begin{eqnarray*}
L = {\alpha}_1 L_{rec}(s,s') + {\alpha}_2 L_{KL} +  {\alpha}_3 L_{rec}(s,G(\bar{z})) + {\alpha}_4 L_{mask}(k, \tilde{k}),
\end{eqnarray*}
where $\bar{z} = \tilde{z}$ if training the topogram-mask encoder, and $\bar{z} = \hat{z}$ when training the topogram-only encoder. Note that ${\alpha}_1 L_{rec}(s,s')$ is the reconstruction loss of the VAE and ${\alpha}_3 L_{rec}(s,G(\bar{z}))$ is the 2D-3D reconstruction loss. It is also possible to train the above model without the shape encoder, i.e. ${\alpha}_1=0$ and ${\alpha}_2=0$. 

In all experiments, we use ${\alpha}_1=50.0$, ${\alpha}_2 = 0.1$, ${\alpha}_3 = 50.0$ and ${\alpha}_4 = 0.0001$ if the mask is provided as input (see Section~\ref{sec:topoCont}) or ${\alpha}_4 = 0$ otherwise (for topogram only approach). All models are trained using Adam optimizer~\cite{adam} with learning rate $0.0001$ for $250$ epochs and batch size of $32$.

\section{Experimental Results and Discussion}
We perform extensive quantitative and qualitative experiments of our method on the difficult tasks of estimating 3D shape of the human liver and predicting its volume. Due to their heterogeneous and diffusive shape, automatic liver segmentation is a very complex problem. Using our method we can accurately estimate the 3D shape of the liver from a 2D topogram image and optionally a 2D mask. We use voxel grids as our base representation, and visualize results using 2D projections or 3D meshes obtained using marching cubes \cite{lorensen1987marching}. 

We investigate the effect of shape context provided by the mask observations by evaluating a baseline where 3D shape is predicted directly from the mask. We also quantitatively compare our method to an adversarial baseline~\cite{wu2016single} approach. 

\subsection{Dataset}
To conduct an experimental evaluation, we collected $2129$ abdominal CT scans (3D volumetric images of the abdomen covering the liver organ) from several different hospital sites. The liver shapes were segmented using volumetric segmentation approach~\cite{yang2017automatic} and topograms and masks are extracted via 2D projection. Examples from the dataset as well as the provided annotations are shown in Figure~\ref{fig:examples-liver}. We use $1554$ scans for training, and $575$ for testing.

We demonstrate several direct applications of our method: three-dimensional shape reconstruction with corresponding two-dimensional liver delineation through projection and organ volume prediction.

\begin{figure}[!th]
\centering
\includegraphics[width=1.0\linewidth]{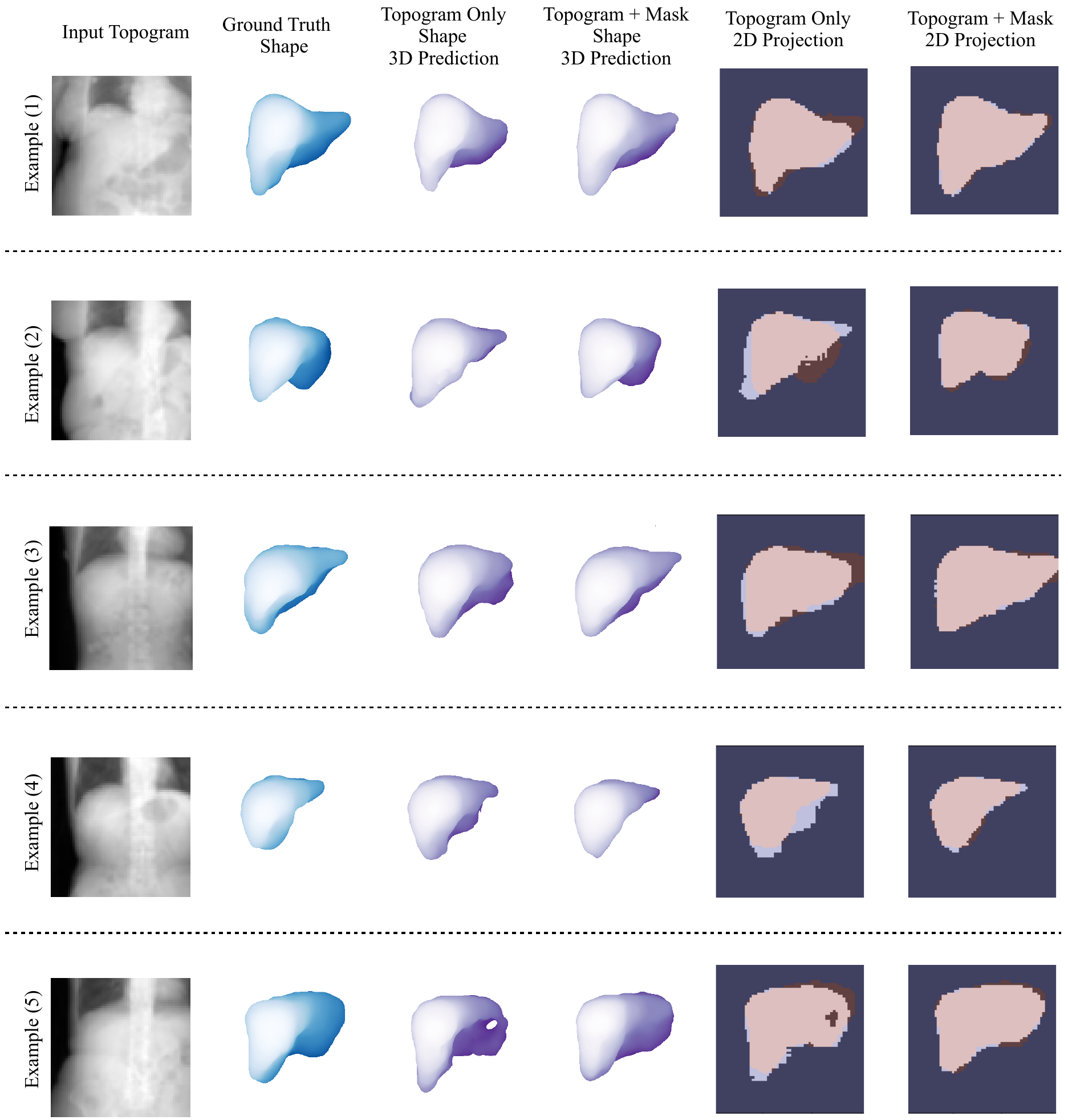}

\caption{Sample examples of 3D reconstruction.  The input topograms are shown in first column, the ground truth shapes are shown in second column, and the predicted shapes are shown in purple (third column - topogram only approach, fourth column - topogram + mask approach). The projected masks of the corresponding approaches, overlaid with the ground truth masks, are shown in the fifth and sixth columns, respectively. The ground truth mask is shown in pink, and the predicted mask is shown in light purple (the predicted background segmentation is dark purple).}
\label{fig:liver-recons}
\end{figure}

\subsection{Organ Shape Reconstruction from Topograms}
Given a learned generative model of liver shapes and an image encoder which estimates a latent space vector given a topogram image (and mask, if given), we predict the 3D liver shape, and project it back onto the topogram image plane to perform two-dimensional delineation. Visually delineating accurate shape from topograms is particularly difficult due to visual ambiguities, such as color contrast and fuzzy boundaries. Our method can predict the three-dimensional shapes automatically from the topogram, and refine the prediction, given a two-dimensional mask annotation.  

\subsubsection{Qualitative Evaluation}
In Figure~\ref{fig:liver-recons}, we visualize the 3D reconstruction results.  The first column is a visualization of the input topogram, the second column is the visualization of the ground truth 3D shape, the third column is the visualization of the result of the topogram-only approach, the fourth column is the visualization of the result of the topogram+mask approach, and the fifth and sixth columns are visualizations of projected masks of the corresponding two approaches, overlaid with the ground truth masks.  Each row corresponds to a different example.

% right lobe % (top left liver part) 
% elongated interior tip (top right liver tip)
% right lobe tip (bottom left liver tip)
Both proposed methods are able to capture significant variation in the observed shapes, such as a prominent dome on the right lobe in Example $1$ and shape of the left lobe in Example $5$. The topogram+mask method is able to convey more topological details compared to the topogram-only method: an elongated interior tip in Examples $1$ and $4$, protrusion off left lobe in Examples $2$ and $3$, and overall topology in Example $5$, where the mask-based method corrects the hole artifact introduced by the topogram-only method. Overall, the surfaces in predictions from the mask-based method are visually closer to the ground truth. 

\begin{figure}[!th]
\centering
\includegraphics[width=0.9\linewidth]{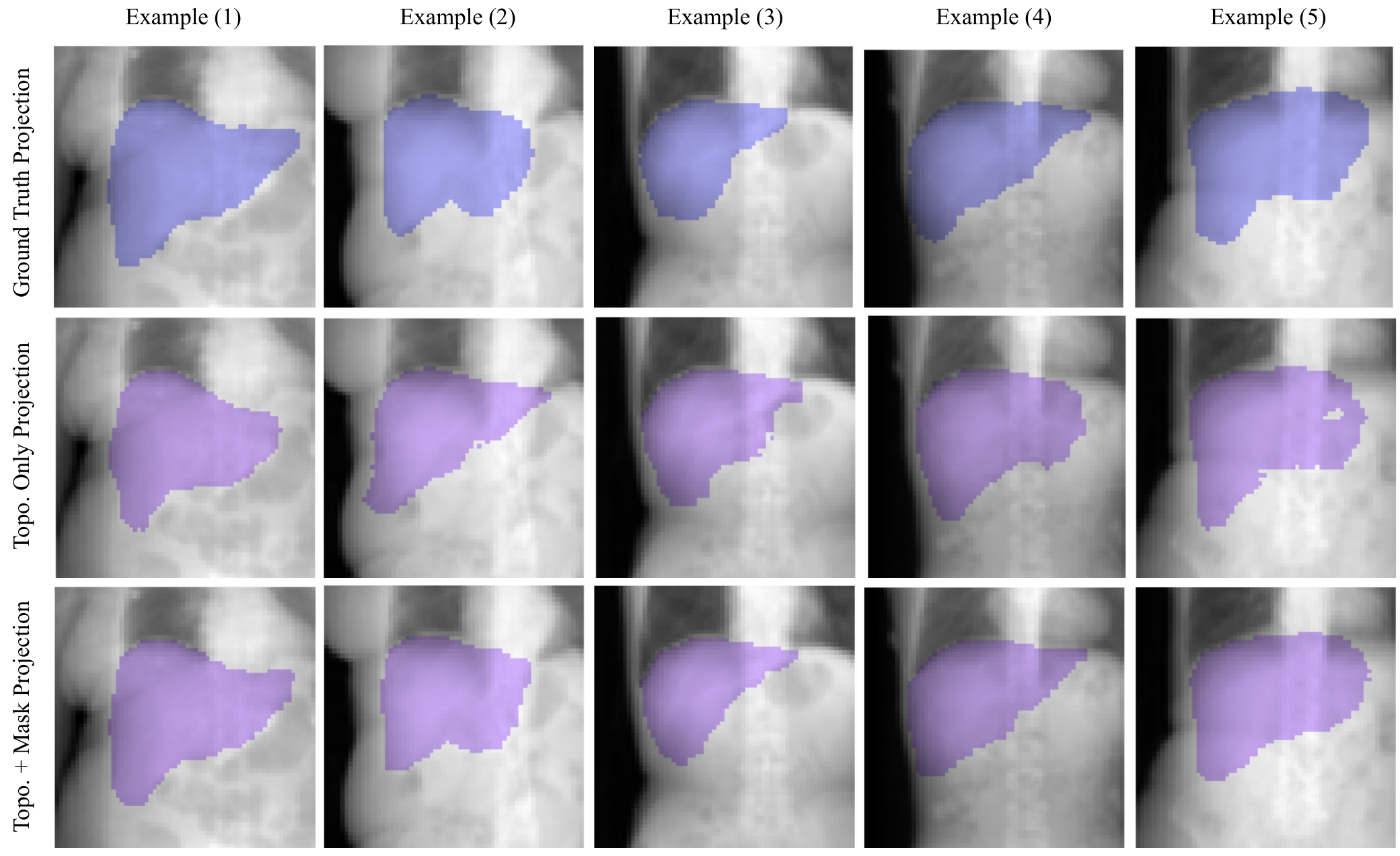}

\caption{Sample 2D projections of the predicted organ shapes. The ground truth projections are shown in first row, the topogram only prediction projections are shown in second row, and the topogram+mask projections are shown in third row. By predicting the 3D shape of the organ, we are also able to generate an accurate 2D segmentation of the input topograms via projection.}
\label{fig:liver-proj}
\end{figure}

We also project the 3D predictions directly on the input topograms (see Figure~\ref{fig:liver-proj}). This allows us to visualize the corresponding inferred 2D segmentation. The shape reconstruction network (in both topogram only and topogram+mask methods) learns to emphasize on characteristic parts of the organ shape, such as the curves in the  right lobe and interior tip.  

\subsubsection{Quantitative Evaluation} \label{sec:quant_eval}
Several metrics can be used to quantitatively compare 3D shape reconstructions (see \cite{christ2017automatic} for details). We provide a quantitative evaluation using two popular volume-based metrics (Intersection over Union (IoU) and Dice coefficients) and a surface-based metric (Hausdorff distance) in Table~\ref{table:liver-3d-shape-prediction-tab}. The topogram+mask approach outperforms the topogram only approach according to all of the metrics, but especially according to Hausdorff distance, which is very sensitive to shape variations such as critical cases of incorrect tip or bulge presence prediction. 

\begin{table}[!h]
\centering
   \begin{tabular}{C{2.0cm} C{2.5cm}C{2.5cm}C{2.5cm}}
%   \begin{tabular}{C{1.3cm} C{1.3cm}C{1.3cm}C{1.3cm} C{1.3cm} C{1.3cm}}
        \hline
        Metric (Mean) & Mask Only       & Topogram Only & Topogram + Mask    \\ \hline
        IOU           &   0.58          & 0.78       & \textbf{0.82}            \\ 
        Dice          &   0.73          & 0.87       & \textbf{0.90}            \\
        Hausdorff     &  28.28          & 7.10       & \textbf{5.00}               \\ \hline
    \end{tabular}
	\caption{Quantitative comparison of the mask only, topogram only and topogram+mask methods on 3D shape reconstruction using volumetric (IoU and Dice metrics) and surface-based metrics (Hausdorff distance).}
	\label{table:liver-3d-shape-prediction-tab}
% \vspace{-1.0cm}
\end{table}

\subsubsection{Shape Context} \label{sec:context}
It is important to investigate whether the provided mask provides too much context, rendering the problem of 3D shape prediction a much easier task. We thus train a mask-only baseline that learns to reconstruct 3D shape directly from mask (no topogram image provided). In Table~\ref{table:liver-3d-shape-prediction-tab}, we compare the performance of this baseline and the two methods that receive the topogram as input. The mask only method is unable to achieve the same quality of results as the topogram-based methods, generating significantly lower mean IoU and Dice errors, and a much larger Hausdorff error. The topogram images contain important information, such as shape layout, that is complementary to the context extracted from masks, and thus both inputs are needed for high quality reconstruction.

\subsection{Volume Calculation}
Of particular interest in the medical community is the automatic volume measurement of main organs. Our method predicts the 3D shape, which we can directly use to measure organ volume. In Table~\ref{table:liver-3d-volume-surface}, we evaluate our proposed approaches on the task of volume prediction. We use the volume of the voxelized 3D segmentation of the liver, obtained from segmentation of the 3D CT, as the ground truth. Given the 3D shape prediction, we measure the predicted volume as the number of voxels in the generated shape (which can be converted to milliliters (mL) using scanning configuration parameters). We report the volume error prediction $V_f = \|V_{pred} - V_{gt}\|/V_{gt}$ where $V_{pred}$ and $V_{gt}$ are the volumes of the predicted and ground truth organs, respectively. 

On average, we are able to predict liver volume to $6\%$ error with the topogram+mask method and to $10\%$ error with the topogram only method. The mask-only based method is unable to predict volume accurately, since it cannot predict the correct 3D topology (see Section~\ref{sec:context}).

\begin{table}[h!]
\vspace{-0.5cm}
\centering
    \begin{tabular}{C{2.5cm} C{2.5cm} C{2.5cm}  C{2.5cm}}
        \hline
        Metric       & Mask Only          & Topogram Only &  Topogram + Mask    \\ \hline
        Volume Error ($V_f$) & 0.34               & 0.10          &  \textbf{0.06}        \\ \hline
    \end{tabular}
	\caption{Mean Volume Error ($V_f$) evaluation and comparison. }
	\label{table:liver-3d-volume-surface}
\vspace{-1.5cm}
\end{table}
\subsection{Comparison to Adversarial Approaches}
We also compare our method to an adversarial baseline (3D VAE-GAN~\cite{wu3dGAN}) which is another commonly used generative modelling approach. We train this baseline with the same architecture and hyperparameters described in \cite{wu3dGAN}. We observe that the discriminator in this baseline would typically encourage more uniform predictions compared to our VAE-based method, thus discouraging generation of more diverse shape topologies. Quantitatively, this method achieves lower quality results than the both VAE-based methods (see Table~\ref{table:adversarial}), especially in surface-based error and volume error due to its tendency to predict an average shape irrespective of the input.

\begin{table}[]
\vspace{-0.5cm}
\begin{tabular}{l|c|ccc|}
\cline{2-5}
\multicolumn{1}{c|}{}                                                                                             & Volume Prediction & \multicolumn{3}{c|}{Shape Reconstruction}                                                      \\ \cline{2-5} 
                                                                                                                  & Volume Error ($V_f$)          & IoU                           & Dice                          & Hausdorff                      \\ \hline
\multicolumn{1}{|l|}{\begin{tabular}[c]{@{}l@{}}Variational Autoencoder (VAE)\\ (without/with mask)\end{tabular}} & 0.10/\textbf{0.06}         & \multicolumn{1}{l}{0.78/\textbf{0.82}} & \multicolumn{1}{l}{0.87/\textbf{0.90}} & \multicolumn{1}{l|}{7.10/\textbf{5.00}} \\ \hline
\multicolumn{1}{|l|}{Adversarial (3D-GAN)~\cite{wu3dGAN}}                                                                        & 0.21              & 0.61                          & 0.75                          & 10.50                          \\ \hline
\multicolumn{1}{|l|}{Performance Difference}                                                                      & 109\% / 250\%     & 22\% / 26\%                   & 14\% / 17\%                   & 48\% / 110\%                   \\ \hline
\end{tabular}
\caption{Comparison of the variational autoencoder (VAE) based approaches to a generative adversarial network (GAN) based approach on volume prediction and shape reconstruction tasks.}
\label{table:adversarial}
\vspace{-1.5cm}
\end{table}

\subsection{Conclusion and Future Work}
3D organ shape reconstruction from topograms is an extremely challenging problem in medical imaging. Among other challenges, it is a difficult problem because the input X-ray images can contain projection artifacts that reconstruction methods need to handle, in addition to predicting the topology of occluded and unseen parts the three-dimensional organ. The core insight of this work is that, despite the visual ambiguities present in this type of imagery, it is possible to predict 3D organ shape directly from topograms. It is also possible to improve the quality of the prediction by providing supplementary two-dimensional shape information in the form of masks. 

This work is only a first step towards performing more accurate and reliable 3D organ shape reconstruction. It would be interesting to investigate the performance of our approach on organs other than liver, such as lung or heart, and explore other types of user inputs and annotations that can improve the reconstruction quality. Also, it would be critical to study why 2D to 3D mapping is possible, and what types of neural networks (in this work we focused on the VAE) are best suited for modelling the shape space and achieving high reconstruction accuracy. Further, categorizing the dataset according to data perturbations, such as fatty liver, tumors, liver disease, age or gender, one should study how these factors affect the performance accuracy. Finally, it would be important to analyze how X-ray can help improve reconstruction accuracy when 3D scans are available and extracting liver shape can be posed as a 3D shape segmentation problem. We hope this work will inspire other approaches that apply generative 3D modelling techniques to reconstructing and predicting organ shapes.

\subsection{Acknowledgements}
We thank Daguang Xu for help with anatomical part labelling and discussions; Thomas Funkhouser, Terrence Chen, Kai Ma, and members of the Princeton Graphics and Vision Group for helpful suggestions; Sungheon Gene Kim, Linda Moy, Krzysztof Geras, and Kyunghyun Cho for discussions on medical applications of the proposed method. This work was supported by Siemens Healthcare and NSF-GRFP.

{\footnotesize
\bibliographystyle{splncs04}
\bibliography{egbib}
}
\end{document}